\documentclass[
]{ceurart}

\usepackage{listings}
\usepackage{indentfirst}
\usepackage{amsmath}
\usepackage[american]{babel}
\usepackage{microtype}
\begin{document}

\copyrightyear{2023}
\copyrightclause{Copyright for this paper by its authors.
  Use permitted under Creative Commons License Attribution 4.0
  International (CC BY 4.0).}

\conference{CLEF 2023: Conference and Labs of the Evaluation Forum, September 18–21, 2023, Thessaloniki, Greece}

\title{Watch out Venomous Snake Species: A Solution to SnakeCLEF2023}



\author[1]{Feiran Hu}[%
email=hufr@njust.edu.cn,
]
\author[1]{Peng Wang}[%
email=wangpeng@njust.edu.cn,
]
\author[1]{Yangyang Li}[%
email=lyylyyi599@njust.edu.cn,
]
\author[1]{Chenlong Duan}[%
email=duancl@njust.edu.cn,
]
\author[1]{Zijian Zhu}[%
email=zhuzj@njust.edu.cn,
]
\author[2]{Fei Wang}[%
email=wangfei@ainnovation.com,
]
\author[2]{Faen Zhang}[%
email=zhangfaen@ainnovation.com,
]
\author[1]{Yong Li}[%
email=yong.li@njust.edu.cn,
]
\cormark[1]
\author[1]{Xiu-Shen Wei}[%
email=weixs.gm@gmail.com,
]
\cormark[1]
\address[1]{School of Computer Science and Engineering, Nanjing University of Science and Technology, Nanjing, China}
\address[2]{Qingdao AInnovation Technology Group Co,. Ltd.}

\cortext[1]{Corresponding author.}

\begin{abstract}
    The SnakeCLEF2023 competition aims to the development of advanced algorithms for snake species identification through the analysis of images and accompanying metadata. This paper presents a method leveraging utilization of both images and metadata. Modern CNN models and strong data augmentation are utilized to learn better representation of images. To relieve the challenge of long-tailed distribution, seesaw loss~\cite{wang2021seesaw} is utilized in our method. We also design a light model to calculate prior probabilities using metadata features extracted from CLIP~\cite{radford2021learning} in post processing stage. Besides, we attach more importance to venomous species by assigning venomous species labels to some examples that model is uncertain about. Our method achieves 91.31\% score of the final metric combined of F1 and other metrics on private leaderboard, which is the 1st place among the participators. The code is available at \url{https://github.com/xiaoxsparraw/CLEF2023}.
\end{abstract}


\maketitle

\section{Introduction}

\noindent Fine-grained visual categorization is a well-established and pivotal challenge within the fields of computer vision and pattern recognition, serving as the cornerstone for a diverse array of real-world applications~\cite{wei2021fine}. The SnakeCLEF2023 competition, co-hosted as an integral part of the LifeCLEF2023 lab within the CLEF2023 conference, and the FGVC10 workshop in conjunction with the esteemed CVPR2023 conference, is geared towards advancing the development of a robust algorithm for snake species identification from images and metadata. This objective holds profound significance in the realm of biodiversity conservation and constitutes a crucial facet of human health preservation. 

\indent In this paper, we introduce a method that addresses the recognition of snake species by leveraging both metadata and images. ConvNeXt-v2~\cite{woo2023convnext} and CLIP~\cite{radford2021learning} are used to extract images features and metadata features separately, and the image features and text features are concatenated to be input of MLP classifier, thus getting better representation of examples and recognition results. Seesaw loss~\cite{wang2021seesaw} are utilized in our method,  thereby alleviating the long-tailed distribution problem. Notably, our proposed method takes into careful consideration the critical real-world need to distinguish venomous and harmless snake species by using the Real-World Weighted Cross-Entropy (RWWCE) loss~\cite{ho2019real} and post-processing, resulting in exemplary performance surpassing that of other solutions presented in this year's competition. Experiments and competition results show that our method is effective in snake species recognition task.

\indent 
The subsequent sections of this paper provide a comprehensive overview of the key aspects. Section~\ref{s_dataset} introduces the competition challenges and datasets, accompanied by the examination of the evaluation metric utilized. Section~\ref{s_method} describes our proposed methodologies, offering a comprehensive and detailed introduction to the techniques. Section~\ref{s_experiments} presents the implementation details, alongside a comprehensive analysis of the principal outcomes achieved. Finally, Section~\ref{s_conclusion} concludes this paper by summarizing the key findings and offering future research directions.

\section{Competition Description\label{s_dataset}}

\noindent Understanding datasets and metrics is an essential requirement for engaging in a machine learning competition. Within this section, we aim to introduce our comprehension of the datasets and provide overview of the evaluation metrics employed by the competition organizers.

\subsection{Challenges of the Competition}

\noindent Past iterations of this competition have witnessed remarkable accomplishments by machine learning models~\cite{picek2020overview, picek2021overview, picek2022overview, bloch2020combination, chamidullin2021deep, zou2022solutions}. To further enhance the competition's practical relevance and address the exigencies faced by developers, scientists, users, and communities, such as addressing post-snakebite incidents, the organizers have imposed more stringent constraints. The ensuing challenges of this year's competition can be summarized as follows:

\begin{itemize}
\item  Fine-grained image recognition: The domain of fine-grained image analysis has long posed a challenging problem within the FGVC workshop, deserving further investigation and study.
\item Utilization of metadata: The incorporation of metadata, particularly pertaining to the geographical distribution of snake species, plays a vital role in their classification. Such metadata is commonly employed by individuals to identify snakes in their daily lives. Hence, utilization of location metadata holds significance and needs careful consideration.
\item Long-tailed distribution: Long-tailed distributions are common in real-world scenarios, and the distribution of snake species is no exception.
\item Identification of venomous and harmless species: The distinction between venomous and harmless snake species is meaningful, as venomous snake bites lead to large number of death each year. Consequently, leveraging deep learning methodologies to address this problem is of paramount urgency.
\item Model size limitation: A strict limitation has been imposed on the model size, constraining it to a maximum of 1GB.
\end{itemize}

\subsection{Dataset}

\noindent The organizers provide a dataset, consisting of 103,404 recorded snake observations, supplemented by 182,261 high-resolution images. These observations encompass a diverse range of 1,784 distinct snake species and have been documented across 214 geographically varied regions.

\begin{figure}
  \centering
  \includegraphics[width=\linewidth]{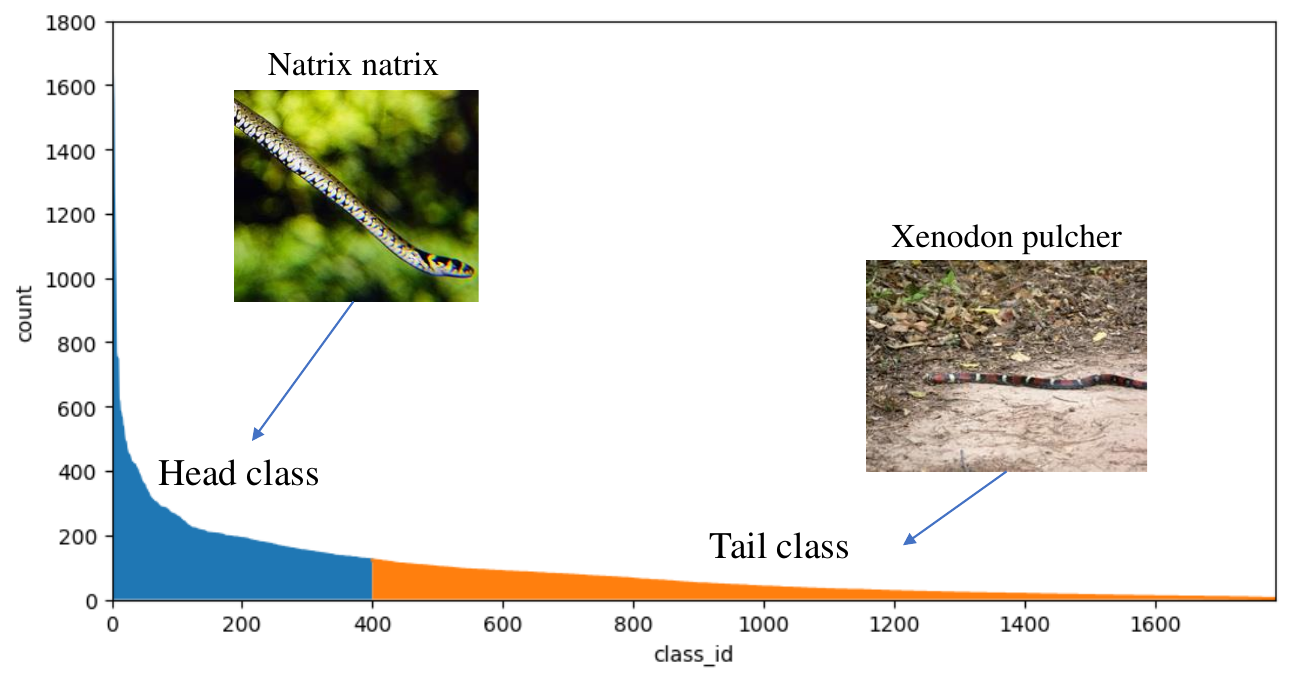}
  \caption{Long-tailed distribution of the SnakeCLEF2023 training dataset. The blue color means head classes, which means most images in the dataset belong to these classes. The orange color means tail classes, which means most classes in the dataset are tail classes.}
  \label{fig:longtail}
\end{figure}

\indent It is worth to note that the provided dataset is in a heavily long-tailed distribution, as shown in Fig.~\ref{fig:longtail}. In this distribution, the most frequently encountered species have 1,262 observations consists of 2,079 accompanying images. However, the least frequently encountered species is captured by a mere 3 observations, showing its exceptional rarity within the dataset.

\subsection{Evaluation Metric}

\noindent In addition to the conventional evaluation metrics of Accuracy (Acc) and Mean F1-Score, this year's competition incorporates a novel evaluation metric, denoted as ``public\_score\_track1'' on the leaderboard. This metric combines the F1-Score with an assessment of the confusion errors related to venomous species. It is calculated as a weighted average, incorporating both the macro F1-score and the weighted accuracy of various types of confusions:
\begin{equation}\label{metric}
M=\frac{w_1 F_1+w_2\left(100-P_1\right)+w_3\left(100-P_2\right)+w_4\left(100-P_3\right)+w_5\left(100-P_4\right)}{\sum_i^5 w_i} \,,
\end{equation}
where $w_1=1.0, w_2=1.0, w_3=2.0, w_4=5.0, w_5=2.0$ are the weights of individual terms. The metric incorporates several percentages, namely $F_1$ representing the macro F1-score, $P_1$ denoting the percentage of harmless species misclassified as another harmless species, $P_2$ indicating the percentage of harmless species misclassified as a venomous species, $P_3$ reflecting the percentage of venomous species misclassified as another harmless species, and $P_4$ representing the percentage of venomous species misclassified as another venomous species.  This metric is bounded below by 0\% and above by 100\%.  The lower bound is attained when all species are misclassified, including misclassification of harmless species as venomous and vice versa. Conversely, if the F1-score reaches 100\%, indicating correct classification of all species, each $P_i$ value must be zero, leading to an overall score of 100\%.

\section{Method\label{s_method}}

\noindent In this section, we shall introduce the methodologies employed to address the task of snake species classification.

\subsection{Data Preprocessing}

\noindent Data preprocessing plays a crucial role in machine learning, as it influences not only the final performance but also the feasibility of problem resolution. Upon obtaining the dataset provided by the competition organizers, several issues emerged. For instance, certain images listed in the metadata CSV file were found to be nonexistent within the corresponding image folders. To address this, we generated a new metadata CSV file by eliminating the affected rows from the original file. Additionally, a subset of images within the dataset was found to be corrupted, potentially due to network transmission or other factors. To mitigate this concern, we utilized OpenCV to read the problematic images and subsequently re-wrote them to the file system, thereby solving the corruption issue.

\indent The SnakeCLEF dataset includes valuable metadata pertaining to the observation locations. Leveraging this location information is of great significance, as certain snake species inhabit geographically confined areas. However, the metadata presents the location in the form of country or region codes, which cannot be directly utilized as inputs for convolutional neural network (CNN) or Vision Transformer (ViT)~\cite{dosovitskiy2020image}. To address this challenge, we employ CLIP~\cite{radford2021learning} to extract location features without engaging in fine-tuning. Subsequently, Principal Component Analysis (PCA)~\cite{mackiewicz1993principal} is employed to reduce the dimension of the resulting feature vectors.

\indent Data augmentation serves as a key technique in computer vision tasks. Within our methodology, we leverage fundamental image augmentation methods from Albumentations~\cite{buslaev2020albumentations}. These methods encompass RandomResizedCrop, Transpose, HorizontalFlip, VerticalFlip, ShiftScaleRotate, RandomBrightnessContrast, PiecewiseAffine, HueSaturationValue, OpticalDistortion, ElasticTransform, Cutout, and GridDistortion. Furthermore, we incorporate data mixing augmentation techniques, such as Mixup~\cite{zhang2017mixup}, CutMix~\cite{yun2019cutmix}, TokenMix~\cite{liu2022tokenmix}, and RandomMix~\cite{liu2022randommix}, during the course of the competition. These data augmentation methods provide strong regularization to models by softening both images and labels, avoiding the model overfitting in training dataset.

\subsection{Model}

\noindent Throughout the competition, we explored various models, including both classical and state-of-the-art architectures, such as Convolutional Neural Networks and Vision Transformers. Models employed during the competition include ResNet~\cite{he2016deep}, VOLO~\cite{yuan2022volo}, ConvNeXt~\cite{liu2022convnet}, BEiT-v2~\cite{peng2022beit}, EVA~\cite{fang2023eva} and ConvNeXt-v2~\cite{woo2023convnext}. The implementation of these models was facilitated using the timm~\cite{rw2019timm} library. In light of the imposed limitations on model parameters and the consideration of the model representation capabilities, we selected ConvNeXt-v2~\cite{woo2023convnext} as the backbone architecture in our final method.

\begin{figure}
  \centering
  \includegraphics[width=\linewidth]{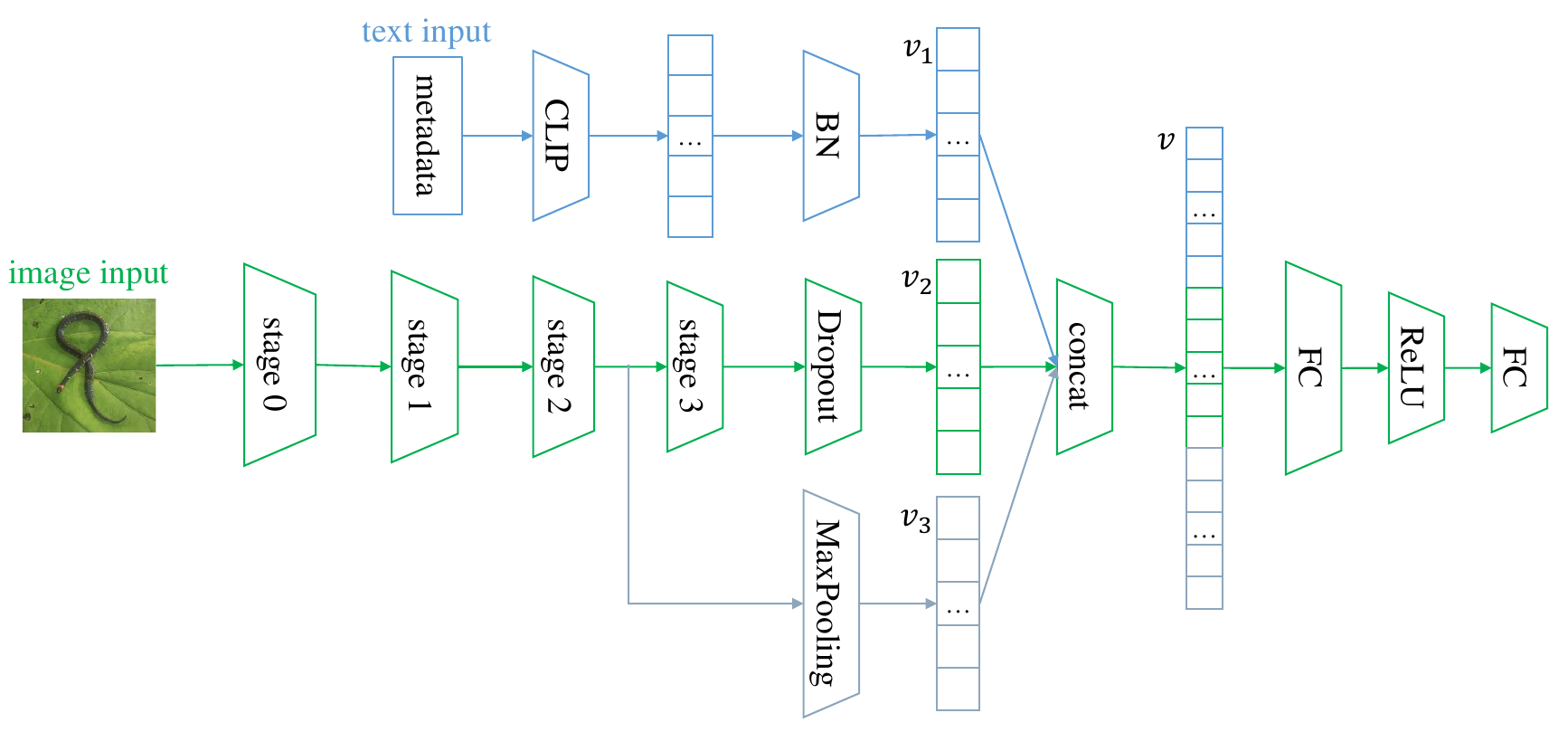}
  \caption{Architecture of our model. Take ConvNeXt-v2~\cite{woo2023convnext} as the backbone, which is made up of 4 stages, feature vector extracted from metadata ($v_1$), original feature vector ($v_2$) and feature vector from middle stage of the backbone ($v_3$) are concatenated to get the final feature vector $v$, a MLP classifier is followed to get the final classification results. }
  \label{fig:architecture}
\end{figure}

\indent 
However, relying solely on the visual backbone is insufficient for effectively addressing the task at hand. Given the availability of metadata in the competition and the inherent challenges associated with fine-grained image classification, it becomes necessary to modify the architecture of the vision model to achieve superior performance. The architectural design of the model employed in our final submission is illustrated in Fig.~\ref{fig:architecture}.

Following the completion of the third stage of ConvNeXt-v2~\cite{woo2023convnext}, the intermediate-level feature map is combined with the high-level image features after the final stage, along with the metadata features. This concatenation process yields a comprehensive representation that captures both the image and metadata information. To mitigate overfitting, we have incorporated MaxPooling~\cite{boureau2010learning}, BatchNorm~\cite{ioffe2015batch}, and Dropout~\cite{hinton2012improving} techniques into our methodology. Once the comprehensive representation is obtained, a classifier comprising two linear layers and ReLU~\cite{nair2010rectified} activation functions follows and generates classification results.

\subsection{Optimization Procedure}

\noindent Addressing long-tailed recognition is another challenge encountered in the competition. To tackle this issue, we extensively explored various techniques implemented in BagofTricks-LT~\cite{zhang2021tricks}. In our final submission, we incorporated the seesaw loss~\cite{wang2021seesaw} as a key component. The seesaw loss formulation can be expressed as follows:

\begin{equation}\label{seesawloss}
\begin{aligned}
& L_{\text {seesaw }}(\mathbf{z})=-\sum_{i=1}^C y_i \log \left(\widehat{\sigma}_i\right)\,, \\
& \text { with } \widehat{\sigma}_i=\frac{e^{z_i}}{\sum_{j \neq i}^C \mathcal{S}_{i j} e^{z_j}+e^{z_i}}\,,
\end{aligned}
\end{equation}

\noindent where $\mathbf{z}$ denotes the output obtained from the fully connected layer, $C$ represents the total number of classes, and $y_i$ corresponds to the one-hot label of the image. The hyper-parameters $\mathcal{S}_{ij}$ are carefully set based on the distribution characteristics inherent in the dataset.

Distinguishing between venomous and non-venomous snake species and the consequential assignment of varying costs to different classification errors are of great importance in this year's challenge, as demonstrated by Eq.~\ref{metric}. In accordance with these requirements, loss function that effectively models the real-world costs associated with mislabeling~\cite{ho2019real} is utilized by us. To align with this objective, we incorporate the Real-World Weighted Cross-Entropy (RWWCE) loss function~\cite{ho2019real} during the final three epochs of training, employing a reduced learning rate.

In addition to the choice of loss functions, the selection of an optimizer and an appropriate learning rate decay strategy are important in the training of our models. For optimization, we adopt the AdamW optimizer~\cite{loshchilov2017fixing}. To enhance convergence speed and overall performance, we implement cosine learning rate decay~\cite{loshchilov2016sgdr} coupled with warmup techniques during the training process. These strategies collectively facilitate more effective and efficient model convergence.

\subsection{Post-processing}
\noindent In this year's challenge, the task requires the solution to accurately identify the venomous nature of snakes, particularly focusing on distinguishing the venomous species, with the limited model capacity. It is challenging but fortunately, the organizers provided a metadata repository, with a particular focus on geographical information. In practical contexts, where reliance solely on visual cues may prove insufficient performance on fine-grained classification, the supplementation of geographical details assumes a crucial role in assisting human experts in making judgment. Thus, the integration of geographical information within the metadata exhibits the potential to enhance the decision-making prowess of classification models.\\
\indent Inspired by~\cite{Aodha_2019_ICCV}, assuming the above-mentioned trained model as $f$, we developed a simple prior model denoted as $g$. This prior model is simple but efficiently, composed of three fully connected layers with non-linear activation function and employed dropout regularization. In the training process of this light model, we adopt the AdamW~\cite{loshchilov2017fixing} optimizer and performed balanced sampling on the training data, to mitigate the impact of the long-tail distribution in the dataset. The objective of this training process was to minimize the following loss function:

\begin{equation}\label{priorloss}
\begin{aligned}
\mathcal{L}_{loc}(\mathbf{x}, \mathbf{r}, \mathbf{O}, y)= & \lambda \log \left(s\left(g(\mathbf{x}) \mathbf{O}_{:, y}\right)\right)+ \sum_{\substack{i=1 \\
i \neq y}}^C \log \left(1-s\left(g(\mathbf{x}) \mathbf{O}_{:, i}\right)\right)+ \\
& \sum_{i=1}^C \log \left(1-s\left(g(\mathbf{r}) \mathbf{O}_{:, i}\right)\right) \,,
\end{aligned}
\end{equation}
where the metadata features extracted from CLIP is denoted as $\mathbf{x}$. $\mathbf{O}$ is the category embedding matrix, where each column is the prototype of different category, pre-computed by our trained model $f$, e.g., ConvNeXt-v2~\cite{woo2023convnext}. Furthermore, $\mathbf{r}$ signifies a uniformly random location data point, and $\lambda$ serves as a hyper-parameter for weighting positive observations. It is important to note that if a category $y$ has been observed at the spatial location $\mathbf{x}$ within the training set, the value of $s\left(g(\mathbf{x}) \mathbf{O}_{:, y}\right)$ should approximate 1. Conversely, if the category has not been observed, the value should approximate 0.\\
\indent During the inference stage, our prior model efficiently calculates the prior class embeddings denoted as $\mathbf{P}$. Utilizing the following equation: 
\begin{equation}\label{predict}
    \mathbf{S^{\prime}} = Softmax(\mathbf{P}) \odot \mathbf{S} ,
\end{equation}
where $\mathbf{S}$ is the prediction score computed by $f$. We derive the final class scores $\mathbf{S^{\prime}}$ by computing the joint probability of predictions from the two models $f$ and $g$. In real-world scenarios, misclassifying a non-venomous snake as venomous carries significant consequences and is deemed unacceptable. To address this concern, we implement a robust post-processing approach. When the predicted confidence of an image $\mathbf{x}$ is relatively low, we analyze its top-5 predictions. If any of these predictions include a venomous class, we classify the image as venomous. This post-processing technique represents a well-considered compromise between precision and recall. Notably, this approach enable us to get the 1st place in the private leaderboard. We firmly believe that this strategy possesses considerable advantages for practical applications.

\section{Experiments\label{s_experiments}}

\noindent In this section, we will introduce our implementation details and main results.

\subsection{Experiment Settings}

\noindent The proposed methodology is developed utilizing the PyTorch framework~\cite{NEURIPS2019_9015}. All models employed in our approach have been pre-trained on the ImageNet dataset~\cite{5206848}, readily available within the timm library~\cite{rw2019timm}. Fine-tuning of these models was conducted across 4 Nvidia RTX3090 GPUs. The initial learning rate was set to $2 \times 10^{-5}$, and the total number of training epochs was set to 15, with the first epoch dedicated to warm-up, employing a learning rate of $2 \times 10^{-7}$. To optimize model training, we utilized the AdamW optimizer~\cite{loshchilov2017fixing} in conjunction with a cosine learning rate scheduler~\cite{loshchilov2016sgdr}, setting the weight decay to $2 \times 10^{-5}$. During inference on the test dataset, we adopted test time augmentation. Furthermore, considering that an observation may consist of multiple images, we adopted a simple averaging approach to obtain a singular prediction for each observation.

\subsection{Main Results}

\noindent In this section, we present our primary findings attained throughout the challenge, as illustrated in Tab.~\ref{tab:results}. The ``Metric'' column within the table corresponds to the public track1 metric featured on the leaderboard.

\begin{table*}
  \caption{Main results of SnakeCLEF.}
  \label{tab:results}
  \begin{tabular}{cccc}
    \toprule
    Backbone & Resolution  & Metric (\%) & Comments\\
    \midrule
    ResNet50~\cite{he2016deep} & $224 \times 224$ & 72.22 & baseline \\
    BEiT-v2-L~\cite{peng2022beit} & $224 \times 224$ & 82.59 & stronger backbone \\
    BEiT-L~\cite{bao2021beit} & $384 \times 384$ & 88.74 & cutmix \\
    EVA-L~\cite{fang2023eva} & $336 \times 336$ & 86.82 & cutmix \\
    Swin-v2-L~\cite{liu2022swin} & $384 \times 384$ & 88.19 & cutmix \\
    VOLO~\cite{yuan2022volo} & $448 \times 448$ & 88.50 & cutmix \\
    ConvNeXt-v2-L~\cite{woo2023convnext} & $384 \times 384$ & 88.98 & seesawloss + randommix \\
    ConvNeXt-v2-L~\cite{woo2023convnext} & $384 \times 384$ & 89.47 & seesawloss + cutmix \\
    ConvNeXt-v2-L~\cite{woo2023convnext} & $512 \times 512$ & 90.86 & seesawloss + cutmix + metadata \\
    ConvNeXt-v2-L~\cite{woo2023convnext} & $512 \times 512$ & 91.98 & \makecell[c]{seesawloss + cutmix \\ + metadata + middle-level feature} \\
    ConvNeXt-v2-L~\cite{woo2023convnext} & $512 \times 512$ & 93.65 & \makecell[c]{seesawloss + cutmix + metadata \\ + middle-level  feature + post-processing} \\
  \bottomrule
\end{tabular}
\end{table*}

As indicated by Tab.~\ref{tab:results}, the model parameters and image resolution hold crucial significance in image recognition tasks, aligning with conventional expectations. An increase in model parameters and image resolution corresponds to improvement in the public leaderboard score. Furthermore, data augmentation plays as a key factor in enhancing the generalization capacity of models. Notably, CutMix~\cite{yun2019cutmix} outperforms alternative data mixing augmentation techniques, such as RandomMix~\cite{liu2022randommix}, based on our experimental observations.

Metadata plays a pivotal role in the recognition of snake species, enabling models to acquire enhanced representations of observations and thereby achieve superior classification results. In our experiments, the utilization of metadata facilitated the acquisition of enriched contextual information, leading to improved model performance. Additionally, the incorporation of the Seesaw loss~\cite{wang2021seesaw} demonstrated notable efficacy in mitigating the challenges posed by long-tailed distributions, surpassing the conventional CrossEntropy loss. Moreover, the integration of middle-level features proved effective in alleviating the complexities associated with fine-grained image recognition, enabling more precise discrimination between similar snake species.

Given that the final evaluation metric takes into account the demands of real-world applications and imposes greater penalties for misclassifying a venomous snake species as harmless compared to misclassifying a harmless species as venomous, we place significant emphasis on post-processing techniques. Specifically, when the model exhibits uncertainty in its predictions for a particular observation, we adopt a cautious approach and classify it as a venomous snake species based on the top-5 predictions. This post-processing strategy has proven highly advantageous, leading to substantial improvements in both the public leaderboard and the private test data performance, as evidenced by Tab.~\ref{tab:results}.

\section{Conclusion\label{s_conclusion}}
\noindent Fine-grained visual analysis holds great practical significance, particularly in accurately discerning the toxicity of snakes within the domain of snake sub-classification. This paper focuses on addressing the snake classification problem by harnessing the valuable metadata present in the dataset for posterior filtering.  Additionally, a robust post-processing technique is employed to facilitate toxicity identification.  These approaches have culminated in our noteworthy achievement of securing the first-place position in the challenge, attaining an impressive overall evaluation score of 91.31\% on the private leaderboard.


\bibliography{snake}

\appendix

\end{document}